%% file: main.tex
\documentclass[twocolumn]{article}

\usepackage[whole]{bxcjkjatype} 

\usepackage[utf8]{inputenc} 
\usepackage[T1]{fontenc}    
\usepackage{hyperref}       
\usepackage{url}            
\usepackage{booktabs}       
\usepackage{amsfonts}       
\usepackage{graphicx}       
\usepackage{nicefrac}       
\usepackage{xcolor}         
\usepackage{tabularx}
\usepackage{appendix}
\usepackage{subcaption}
\usepackage{listings}
\usepackage{float}

\newcommand{\figref}[1]{Figure~\ref{#1}}
\newcommand{\tabref}[1]{Table~\ref{#1}}

\usepackage{tcolorbox}
\tcbuselibrary{listings,skins,breakable}
\tcbset{
  myprompt/.style={
    colback=white, colframe=black,
    colbacktitle=black, coltitle=white,
    fonttitle=\bfseries,
    before upper={\scriptsize\selectfont},
    boxrule=0.5pt,   
    arc=2mm,         
    left=2mm, right=2mm, top=1mm, bottom=1mm, 
    breakable
  }
}

\newtcolorbox{promptbox}[1][]{myprompt, title={#1}}

\tcbset{
  myExecLog/.style={
    colback=white, colframe=gray,
    fonttitle=\bfseries,
    before upper={\scriptsize\selectfont},
    boxrule=0.5pt,
    arc=2mm,
    left=2mm, right=2mm, top=1mm, bottom=1mm,
    enhanced, breakable
  }
}

\newtcolorbox{execbox}[1][]{myExecLog, title={#1}}

\newcommand{\logdivider}{\par\medskip\noindent\dotfill\par\medskip}

\newcommand{\logcode}[1]{%
  {\ttfamily{#1}}%
}

\newcommand{\logsection}[1]{\textbf{#1}\par\smallskip}



\title{
    \vspace{-2.5cm}
EHR-MCP: Real-world Evaluation of Clinical Information Retrieval by Large Language Models via Model Context Protocol}
\date{}
\usepackage{authblk}

\author[1]{Kanato~Masayoshi\thanks{kanato.masayoshi@keio.jp}}
\author[1]{Masahiro~Hashimoto}
\author[1]{Ryoichi~Yokoyama}
\author[1]{Naoki~Toda}
\author[2]{Yoshifumi~Uwamino}
\author[2]{Shogo~Fukuda}
\author[2]{Ho~Namkoong}
\author[1]{Masahiro~Jinzaki}

\affil[1]{\small Department of Radiology, Keio University Hospital, Tokyo, Japan}
\affil[2]{\small Division of Infectious Diseases and Infection Control, Keio University Hospital, Tokyo, Japan}

\begin{document}

\maketitle

\input{section/abstract}

\newpage

\let\thefootnote\relax\footnotetext{This preprint reports ongoing research with preliminary results and has not been peer-reviewed. The results should be interpreted with caution.}

\input{main_text}


{
\scriptsize
\bibliographystyle{IEEEtran}
\bibliography{Paperpile}
}

\clearpage

\appendix
\setcounter{table}{0}
\renewcommand{\thetable}{S\arabic{table}}
\setcounter{figure}{0}
\renewcommand{\thefigure}{S\arabic{figure}}

\input{supplementary}


\newpage
\end{document}

%% file: section/abstract.tex
\begin{abstract}
\small
\noindent\textbf{Background}  
Large language models (LLMs) show promise in medicine, but their deployment in hospitals is limited by restricted access to electronic health record (EHR) systems. The Model Context Protocol (MCP) enables integration between LLMs and external tools.\par
\noindent\textbf{Objective}  
To evaluate whether an LLM connected to an EHR database via MCP can autonomously retrieve clinically relevant information in a real hospital setting.\par
\noindent\textbf{Methods}  
We developed \textbf{EHR-MCP}, a framework of custom MCP tools integrated with the hospital EHR database, and used GPT-4.1 through a LangGraph ReAct agent to interact with it. Six tasks were tested, derived from use cases of the infection control team (ICT). Eight patients discussed at ICT conferences were retrospectively analyzed. Agreement with physician-generated gold standards was measured.\par

\noindent\textbf{Results}  
The LLM consistently selected and executed the correct MCP tools. Except for two tasks, all tasks achieved near-perfect accuracy. Performance was lower in the complex task requiring time-dependent calculations. Most errors arose from incorrect arguments or misinterpretation of tool results. Responses from EHR-MCP were reliable, though long and repetitive data risked exceeding the context window.\par

\noindent\textbf{Conclusions}  
LLMs can retrieve clinical data from an EHR via MCP tools in a real hospital setting, achieving near-perfect performance in simple tasks while highlighting challenges in complex ones. EHR-MCP provides an infrastructure for secure, consistent data access and may serve as a foundation for hospital AI agents. Future work should extend beyond retrieval to reasoning, generation, and clinical impact assessment, paving the way for effective integration of generative AI into clinical practice. \par
\end{abstract}

%% file: main_text.tex
\begin{figure*}[htbp]
  \centering
  \includegraphics[width=0.8\linewidth]{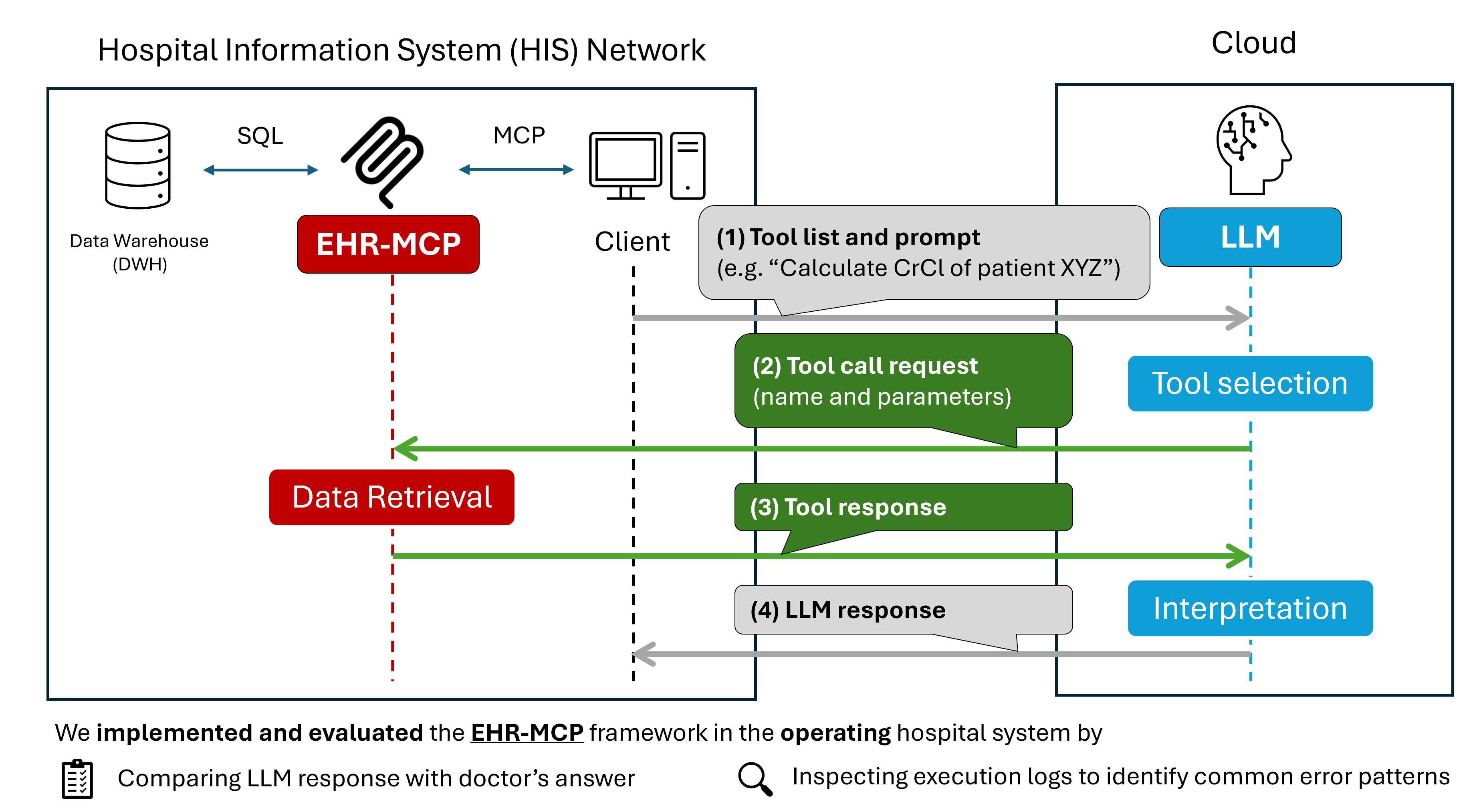}
  \caption{Overview of the EHR-MCP framework.\\
The EHR-MCP framework allows large language models (LLMs) to access hospital electronic health record (EHR) data via the Model Context Protocol (MCP). The LLM client first lists available MCP tools and sends a prompt to the cloud-based model (GPT-4.1). The LLM then selects an appropriate tool and issues a call. MCP retrieves and formats the requested clinical data from the hospital information system (HIS) and returns it to the LLM. The model interprets the response and may issue additional calls until it generates a final answer to the initial prompt.
}
  \label{fig:overview}
\end{figure*}

\section{Introduction}
\subsection{Language models in medicine}
Large language models (LLMs) are generative AI systems trained on vast corpora of text, capable of rapidly comprehending complex inputs and generating coherent responses. Their development has progressed rapidly across diverse domains, including business, science, and software engineering. In medicine, numerous studies have reported potential clinical applications of LLMs\cite{Thirunavukarasu2023-eu}. Examples include generating clinical documents to reduce the burden\cite{Liu2024-sm}, simplifying medical records into patient-friendly language to enhance understanding\cite{Schmiedmayer2024-ry}, and even outperforming physicians in clinical reasoning\cite{McDuff2025-vq, Cabral2024-le}. While LLMs demonstrate strong reasoning and generative performance in benchmark tasks, several challenges remain before they can be applied effectively in real-world clinical environments.

\subsection{The challenge of prompting}
A major obstacle is how to provide LLMs with the breadth and depth of medical information necessary for reliable clinical reasoning. Because LLMs cannot access information beyond what is explicitly supplied, missing data can markedly degrade their performance. Moreover, lacking robust metacognitive capabilities, LLMs may fail to recognize when essential information is absent, and instead generate plausible but incorrect outputs with unwarranted confidence\cite{Griot2025-fi}, a phenomenon known as hallucination\cite{Huang2023-os}.

Extracting clinically relevant information from hospital electronic health record (EHR) systems is itself a nontrivial task for both humans and LLMs. This difficulty arises from the complexity of hospital information systems (HIS), in which multiple subsystems operate concurrently and patient data are dispersed across platforms. For instance, at Infection Control Team (ICT) meetings, assessing the appropriate use of antibiotics requires reviewing patient vital signs, lab results, microbiological culture and susceptibility data, and prior antibiotic exposure. These tasks demand the integration of information from multiple systems. Such processes impose a substantial burden, whether performed by clinicians or delegated to LLMs.

Most existing studies on medical LLMs have evaluated their generative capabilities under the assumption that all requisite information is already curated and supplied in the prompt. This overlooks the practical challenge of information retrieval. In practice, if clinicians must manually collect and format relevant data for each query, the reduction in workload will be minimal. We argue that the most effective paradigm for clinical AI lies in mirroring clinical reasoning: dynamically selecting and applying appropriate tools to iteratively gather and integrate task-specific information.

\subsection{AI agents}
In recent years, the concept of the "AI agent" has emerged, referring to systems in which LLMs not only process text but also autonomously invoke external tools to retrieve information and perform actions. AI agents alternate between reasoning and acting, selecting appropriate tools to address user queries\cite{Yao2022-gn}. For example, when asked about antimicrobial selection, such an agent could automatically gather and integrate data from clinical notes, laboratory results, microbiological cultures, and medication records. Beyond text generation, the deployment of AI agents requires additional capabilities such as reasoning, memory, planning, and reflection, for which decomposition-based evaluation methods have been proposed\cite{Chen2023-fd}.

As a framework for coordinating LLMs with external tools, the Model Context Protocol (MCP), introduced by Anthropic, is rapidly becoming the de facto standard\cite{Anthropic2024-vh}. MCP provides a unified interface that reduces the rising costs of tool-specific integration and is now being adopted by multiple LLM providers.

Besides using external tools, another strategy for automatically extracting relevant information is retrieval-augmented generation (RAG)\cite{Lewis2020-nv}. RAG encodes documents into a vector space, retrieves semantically related passages, and uses them in generation. While effective for unstructured text such as clinical notes, RAG is less suitable for structured numerical data, such as laboratory values, because embedding models often fail to preserve quantitative magnitudes and unit distinctions\cite{Shah2023-br}. Moreover, this approach differs fundamentally from how clinicians collect information, potentially limiting its compatibility with human–AI collaboration. Of note, RAG and AI agents are not mutually exclusive; in practice, one can use both RAG and AI agents\cite{Singh2025-cy}.

\subsection{Related work}

Several studies have explored linking LLMs to EHRs through tool integration, similar to the present work. For instance, Ehtesham et al.\cite{Ehtesham2025-kx} developed a smartphone-based application linked to the EHR, in which an AI chatbot interacted with patients and providers to support clinical decisions. This approach showed potential but lacked quantitative evaluation. MedAgentBench \cite{Jiang2025-dy} provides a more systematic framework, incorporating both retrieval and action-oriented tasks, achieving high accuracy in information retrieval, but its experiments were restricted to simulated environments rather than real-world EHRs. EHRAgent \cite{Shi2024-ow} adopts an extensible strategy by enabling LLMs to generate SQL queries, yet its retrieval success rate remains low, posing substantial challenges for clinical adoption.

In contrast, our study conducted quantitative evaluation. We implemented commonly used clinical information retrieval tasks as deterministic tools, thereby eliminating procedural errors and demonstrating high accuracy. Furthermore, whereas most prior studies have focused only on English-language clinical settings, investigations in non-English contexts remain scarce; our study addresses this gap.

\subsection{Objectives}
The objective of this study is to evaluate whether an LLM can autonomously retrieve and apply clinically relevant information using MCP tools integrated with EHR (\textbf{EHR-MCP}) in a real-world hospital environment. Specifically, we aim to assess the accuracy of MCP-based information retrieval in supporting ICT activities.

The main contributions of this study are as follows:
\begin{itemize}
  \item \textbf{Quantitative evaluation of tool-use capability}: We quantitatively assessed the LLM’s capability to invoke external tools, distinguishing this function from text generation and reasoning.
  \item \textbf{Real-world integration of MCP}: We implemented and deployed EHR-MCP within an operational hospital system, allowing real-world deployment and evaluation while ensuring adaptability to diverse clinical research settings.
  \item \textbf{Cross-linguistic validation}: By validating performance in both English and Japanese, we provide evidence that MCP tools can support medical AI agents across varied clinical contexts worldwide.
\end{itemize}

\section{Methods}
\subsection{Ethics statement}
This study was approved by the Ethics Committee of Keio University School of Medicine (Approval No. 20241132) and conducted in accordance with the Declaration of Helsinki and the Japanese "Ethical Guidelines for Medical and Biological Research Involving Human Subjects." Because this was a retrospective study with no direct impact on patients, informed consent was waived under an opt-out policy. All interactions with the LLM were performed through a secure VPN under contractual agreements ensuring that input data would not be used for model training.

\subsection{System architecture}

\begin{figure*}[htbp]
  \centering
  \begin{subfigure}{0.45\textwidth}
    \centering
    \includegraphics[height=4cm]{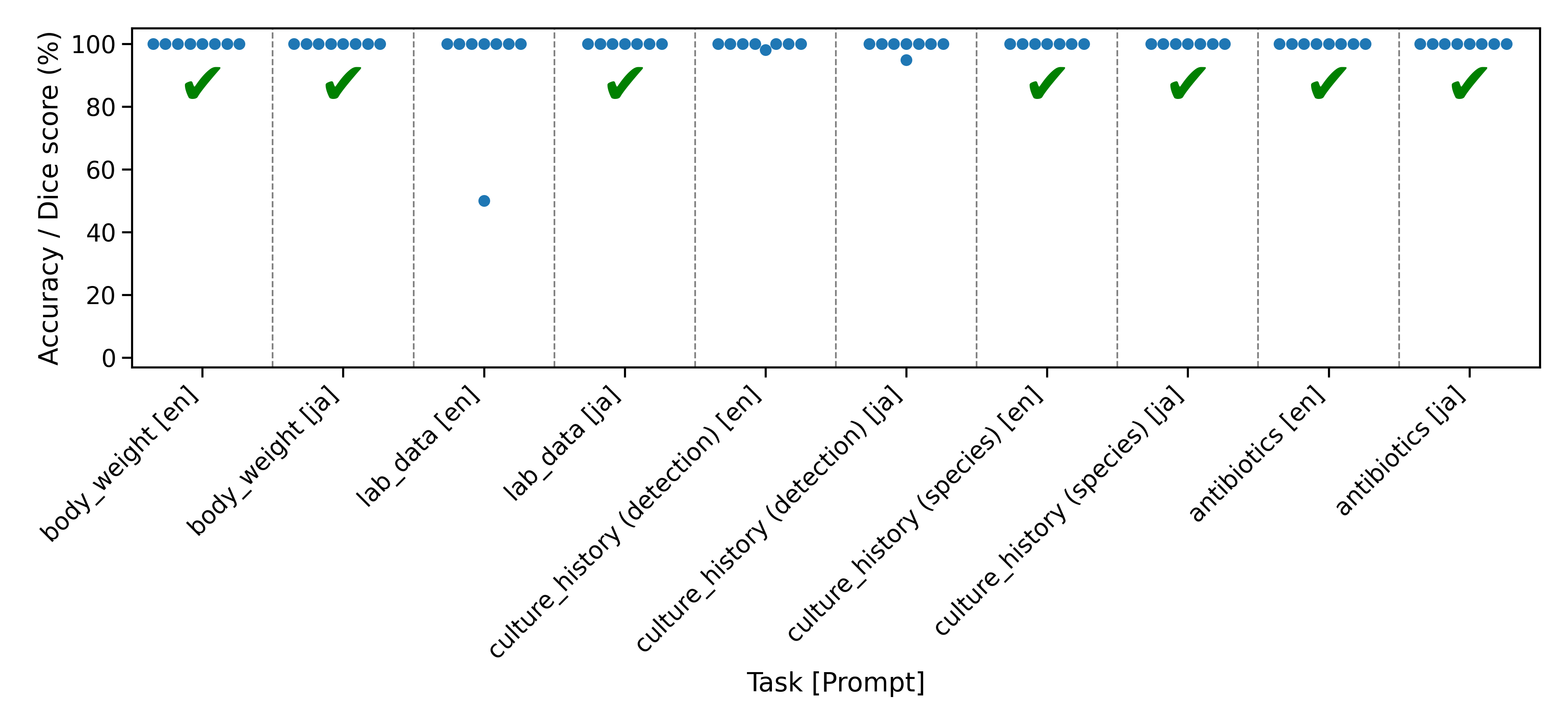}
    \caption{}
    \label{fig:sub1}
  \end{subfigure}
  \hfill
  \begin{subfigure}{0.45\textwidth}
    \centering
    \includegraphics[height=4cm]{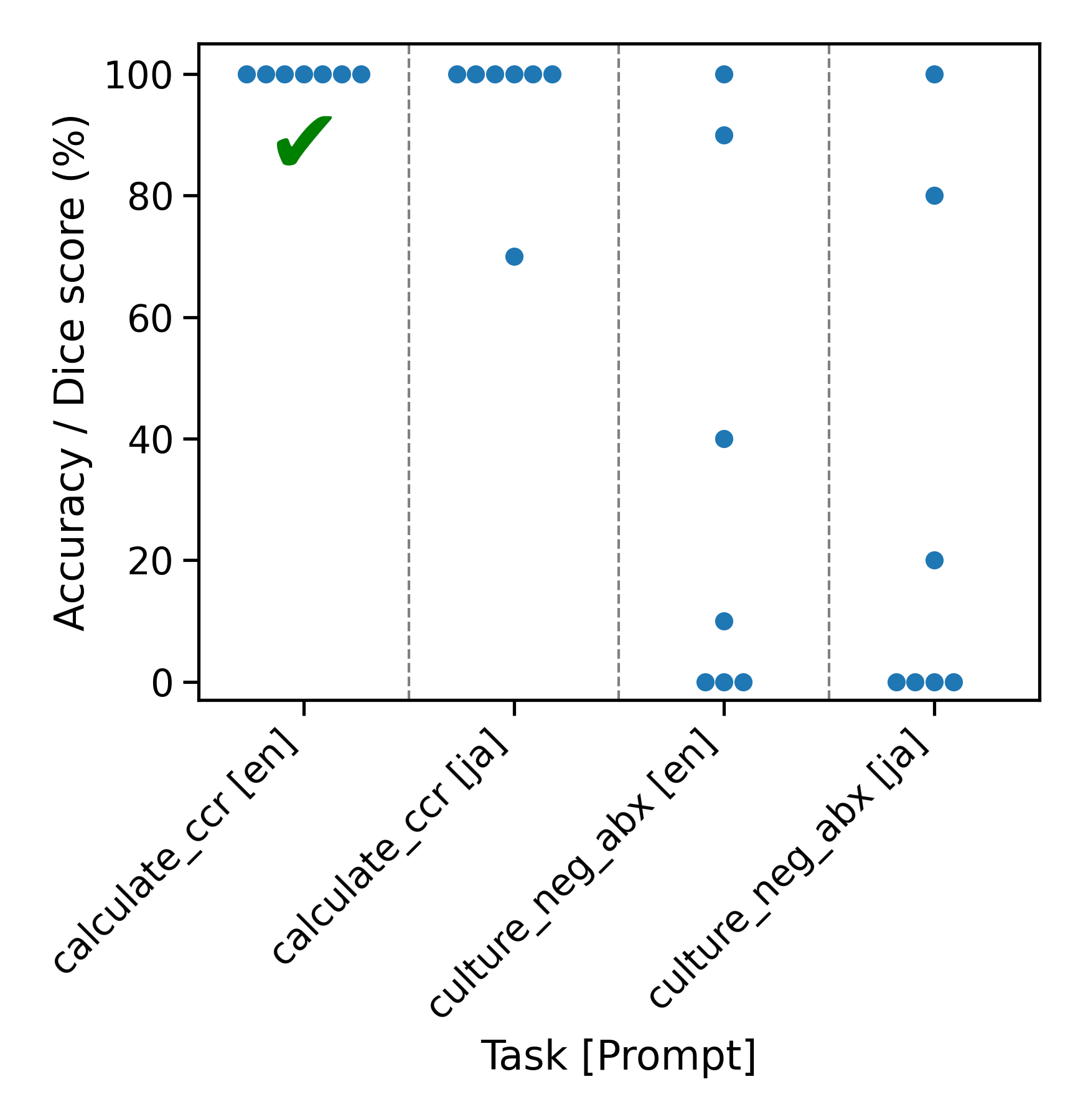}
    \caption{}
    \label{fig:sub2}
  \end{subfigure}
  \caption{Performance (accuracy or Dice score) for each task. (a) Simple tasks, (b) Complex tasks.\\
Each dot represents a patient. For example, in the \texttt{lab\_data} task, seven patients achieved correct retrievals in all 10 trials, while one patient succeeded in only 5 of 10. A green check mark denotes tasks that reached 100\% accuracy across all patients and trials.
}
  \label{fig:result}
\end{figure*}

\figref{fig:overview} provides an overview of the EHR-MCP framework. For the detailed system architecture, see \figref{fig:system_architecture}. Data from the hospital’s electronic health record (EHR) system were synchronized every 24 hours with the in-hospital data warehouse (DWH). The hospital network was securely connected to the Azure OpenAI Service via a VPN through LiteLLM\cite{Unknown2025-ui}, an in-hospital proxy for LLM connections, and GPT-4.1 was employed. A LangGraph ReAct agent\cite{Unknown2025-dx} was used, with structured outputs enforcing JSON-formatted responses. MCP tools were implemented in Python using FastMCP\cite{Lowin2025-ip} (\tabref{tab:mcp-tools}), and data retrieval from the DWH was performed through SQL queries.

\input{tables/mcp-tools}

\subsection{Dataset}

From cases discussed at ICT meetings in 2024, we identified eight patients who had received vancomycin for MRSA bacteremia. To ensure clinical relevance and applicability in ICT, we defined six tasks (\tabref{tab:tasks}). These were categorized into two levels: Simple tasks (\texttt{culture\_history}, \texttt{body\_weight}, \texttt{lab\_data}, \texttt{antibiotics}), which could be completed with a single tool call, and Complex tasks (\texttt{calculate\_ccr}, \texttt{culture\_negative}), which required multi-step tool use and reasoning. Depending on the task, the data retrieval window was set either to the day of intervention initiation or to a one-month period before and after the intervention. Gold-standard answers for each task were prepared by physicians, who manually extracted and formatted the required information from the EHR.

For each task, prompts were prepared in both English and Japanese. They included placeholders for patient ID and the data retrieval period, and examples of the required output format were provided in JSON. Because access restrictions prevented automated retrieval of sex and age that were required for calculating creatinine clearance, these values were manually extracted and embedded into the prompts as placeholders. Details of each task, including the actual prompts, are provided in the Supplementary.

\input{tables/tasks}

\subsection{Evaluation}

The primary evaluation metric was accuracy, defined as the proportion of tasks in which the tool call succeeded and the final output exactly matched the gold standard. For list-based tasks (\texttt{culture\_history} and \texttt{antibiotics}), agreement was measured using the Dice coefficient. Each prompt was executed 10 times to assess stability and reproducibility, and the mean performance across runs was reported.

For error analysis, execution logs were reviewed. These logs included the initial prompt, the invoked tool and its arguments, the tool responses, and the final output. Errors were manually classified into four categories:

\begin{itemize}
  \setlength{\parskip}{0cm}
  \setlength{\itemsep}{0cm}
  \item {Tool Invocation Error:} Failure in invoking the appropriate MCP tool.
  \item {Argument Error:} Incorrect or inappropriate arguments passed to the tool.
  \item {Interpretation Error:} Misinterpretation of the tool response by the LLM.
  \item {Output Format Error:} Output not conforming to the specified JSON schema.
\end{itemize}

\section{Results}

Eight cases were evaluated across six tasks. In the \texttt{calculate\_ccr} task, one patient undergoing dialysis was excluded, as dosing regimens for dialysis patients are generally determined independently of creatinine clearance. In the \texttt{culture\_neg\_abx} task, one case was excluded because no vancomycin administration occurred during the data retrieval period.

Task performance is summarized in \figref{fig:result}. Correct outputs were obtained in all cases for \texttt{body\_weight}, \texttt{culture\_history\_species}, and \texttt{antibiotics}. In \texttt{lab\_data}, one patient yielded incorrect results in 5 of 10 runs. In \texttt{culture\_history\_detection}, one case produced outputs that inappropriately included non-blood culture results.

Among the complex tasks, \texttt{calculate\_ccr} achieved correct results in all but one case, whereas in \texttt{culture\_neg\_abx}, some cases produced incorrect outputs in all 10 runs.

Analysis of execution logs showed that the LLM consistently invoked the expected tools with appropriate arguments across all tasks. No Tool Invocation Errors or Output Format Errors were observed. Because EHR-MCP operates deterministically, correct arguments always produced correct responses.

By contrast, errors arose from Argument Errors and Interpretation Errors. For example, in \texttt{lab\_data} and \texttt{calculate\_ccr}, the model sometimes failed to restrict retrieval to the most recent results, instead using an overly broad range that exceeded the record limit, or conversely, narrowing the retrieval strictly to the intervention day and omitting relevant information. As examples of misinterpretation, \texttt{culture\_history} occasionally included non-blood culture results, and in \texttt{culture\_neg\_abx}, the model misinterpreted tool outputs in the multi-step process, leading to errors in identifying negative culture dates, antibiotic discontinuation, and the day counts between them.

\section{Discussion}

\subsection{Principal findings}
This study demonstrated that LLMs can effectively use MCP-based tools integrated with an EHR to autonomously retrieve clinically relevant information for infectious disease management. Simple tasks requiring a single tool call achieved near-perfect accuracy, whereas errors were more frequent in the \texttt{culture\_neg\_abx} task, highlighting challenges that must be addressed for practical deployment of MCP in clinical settings.

\subsection{Error analysis}
Tool Invocation Errors were not observed, indicating that GPT-4.1 consistently selected appropriate tools through MCP. Similarly, structured outputs prevented Output Format Errors, and no functional failures occurred in EHR-MCP execution.

Most errors arose from Argument Errors and Interpretation Errors. The former typically involved inappropriate specification of the data retrieval window when not explicitly defined. The latter occurred when MCP responses were lengthy or complex, making it difficult for the LLM to extract the intended information. For example, errors were observed when restricting retrieval to blood culture results or when identifying the most recent entry in an antibiotic administration history. These findings indicate that failures mainly occurred at the stages of argument specification and response interpretation. The adverse effects of such errors might be mitigated by interface design strategies informed by human–computer interaction research. For example, increasing the transparency of tool inputs and raw outputs could facilitate human verification and reduce the likelihood of erroneous information entering clinical decision-making.


\subsection{Implementation challenges}
Although MCP tools executed without errors, several implementation issues emerged, highlighting opportunities to improve integration with LLMs. A major challenge is the length of MCP responses: repeated records such as infusions and prescriptions increase with the retrieval window, occasionally exceeding the context capacity of GPT-4. This problem was mitigated after transitioning to GPT-4.1 with an extended context window, but overly long inputs remain undesirable, as they raise API costs and may degrade response quality\cite{Liu2023-zl}. To mitigate this, retrieval windows should ideally be explicitly defined in the prompt. However, in tasks such as \texttt{lab\_data}, the LLM must infer the period to obtain the most recent data, which can lead to errors. Streamlining MCP responses or using auxiliary agents to condense prescription and infusion records within a multi-agent framework may help address these challenges.

Hallucinations were also observed when MCP tools were unavailable or when required information was missing. In some cases, the LLM recognized such failures and returned error messages in JSON format; in others, it produced fabricated outputs. For example, when antibiotic data were unavailable, the model incorrectly reported “CTRX” and “VCM” from the prompt examples as administered (see Supplementary for the execution log). Adding explicit error-handling instructions to prompts may reduce such errors, but anticipating all possible failure scenarios is difficult. Unforeseen failures may still result in fabricated outputs being presented without user awareness.

\subsection{Limitations and future directions}
This study was conducted at a single institution, which limits generalizability. Although the EHR-MCP implementation should be adaptable across institutions by accommodating local system differences at the MCP layer, this has not yet been empirically demonstrated. The sample size and number of tasks were also limited. Future work should examine more diverse and complex tasks with expanded error analysis.

Moreover, this study focused exclusively on tool use capability, while advanced reasoning and generation were beyond its scope. Tool use is foundational for clinical AI agents, but effective deployment will also require robust reasoning and domain knowledge. Just as proficiency in EHR navigation alone does not make a clinician competent without diagnostic reasoning and documentation skills, accurate tool use alone is insufficient. Having established high tool-use accuracy in real-world settings, future research should expand to comprehensive evaluations encompassing reasoning and generation, tailored to specific specialties and diseases. In particular, we plan to investigate whether MCP-retrieved information can support decision-making in ICT.

\subsection{Towards a hospital-wide AI agent infrastructure}
The EHR-MCP implemented in this study represents a reusable AI infrastructure to support a wide range of LLM-based clinical applications. Our ultimate goal extends beyond evaluating tool use capability: by embedding EHR-MCP into hospital operations, we aim to enable rapid, cross-departmental deployment of generative AI projects, thereby accelerating both research and clinical utilization.

Demonstrating the clinical utility of generative AI will require rigorous studies in real-world settings. While many current medical LLM studies rely on retrospective analyses of case series or datasets, future research must demonstrate practical benefits in real-world settings, such as improving patient outcomes and reducing workload. Such studies will likely need to be tailored to specific specialties and diseases; yet, certain challenges will remain common across domains, including secure use of external LLMs under strict privacy protections and efficient retrieval of EHR data. By establishing a shared infrastructure to address these common needs, hospitals can guarantee reliable data access and security, eliminate redundant efforts, and allow research groups to concentrate on domain-specific studies.

\section{Conclusion}
This study demonstrated that LLMs can effectively use MCP-based tools integrated with an EHR to autonomously retrieve clinically relevant information for infectious disease management. By quantitatively evaluating tool use in real-world settings, we showed that simple retrieval tasks can be performed with near-perfect accuracy, while complex tasks revealed challenges related to argument specification and interpretation.

EHR-MCP represents a reusable institutional infrastructure that ensures consistent and secure access to clinical data, thereby reducing redundancy and enabling research groups to focus on advancing clinical and scientific investigations. Building on the high accuracy of tool use, future work should extend evaluations to reasoning and generation, as well as prospective studies assessing the clinical impact of LLM-based systems on outcomes and workflow efficiency. Through such efforts, hospital-wide AI agent infrastructures may help accelerate the safe and effective integration of generative AI into clinical practice.

%% file: tables/mcp-tools.tex
\begin{table*}[htbp]
  \caption{Implemented MCP Tools}
  \label{tab:mcp-tools}
  \centering
  \begin{tabularx}{\textwidth}{l l X}
    \toprule
    \textbf{Tool Name} & \textbf{Parameters} & \textbf{Description} \\
    \midrule
    \texttt{patient\_basic\_info} & \texttt{patient\_id} & Retrieves basic patient information (sex, date of birth, allergies, and latest anthropometric data). \\
    \addlinespace
    \texttt{lab\_results} & \texttt{patient\_id}, \texttt{start\_date}, \texttt{end\_date} & Retrieves clinical laboratory data (e.g., blood and urine tests) within the specified period. \\
    \addlinespace
    \texttt{bacteria\_results} & \texttt{patient\_id}, \texttt{start\_date}, \texttt{end\_date} & Retrieves bacterial culture results (e.g., blood, sputum, urine) within the specified period, including antimicrobial susceptibility. \\
    \addlinespace
    \texttt{antibiotics\_treatment} & \texttt{patient\_id}, \texttt{start\_date}, \texttt{end\_date} & Retrieves antibiotic treatment records (oral and intravenous) within the specified period, limited to predefined antibiotics. \\
    \addlinespace
    \texttt{calculate\_cockcroft\_gault} & \texttt{age}, \texttt{sex}, \texttt{weight}, \texttt{serum\_creatinine} & Calculates creatinine clearance using the Cockcroft–Gault equation based on the given parameters. \\
    \bottomrule
  \end{tabularx}
\end{table*}


%% file: tables/tasks.tex
\begin{table*}[htbp]
  \caption{Tasks and Expected Tool Calls}
  \label{tab:tasks}
  \centering
  \begin{tabularx}{\textwidth}{l l X X}
    \toprule
    \textbf{Difficulty} & \textbf{Task Name} & \textbf{Description} & \textbf{Expected Tool Calls} \\
    \midrule
    Simple & \texttt{body\_weight} & Retrieve the most recent body weight. & \texttt{patient\_basic\_info} \\
           & \texttt{lab\_data} & Retrieve the white blood cell (WBC) count closest to the intervention start date. & \texttt{lab\_results} \\
           & \texttt{culture\_history} & Retrieve the dates and bacterial species of blood cultures within one month before and after the intervention. & \texttt{bacteria\_results} \\
           & \texttt{antibiotics} & List intravenous and oral antibiotics being administered at the intervention start (restricted to predefined list). & \texttt{antibiotics\_treatment} \\
    \midrule
    Complex & \texttt{calculate\_ccr} & Calculate the creatinine clearance closest to the intervention start date. & \texttt{patient\_basic\_info}, \texttt{lab\_results}, \texttt{calculate\_cockcroft\_gault} \\
            & \texttt{culture\_neg\_abx} & Count the number of days vancomycin was administered after blood cultures became negative. & \texttt{bacteria\_results}, \texttt{antibiotics\_treatment} \\
    \bottomrule
  \end{tabularx}
\end{table*}

%% file: supplementary.tex
\section*{Supplementary}
\input{section/technical_details}
\input{section/tasks_and_prompts}
\input{section/execution_logs}

%% file: section/technical_details.tex
\section{Technical Details}
This section provides the technical details necessary to reproduce our framework. See \tabref{tab:library-versions} for the versions of the major libraries used.

\begin{figure*}[htbp]
  \centering
  \includegraphics[width=0.8\linewidth]{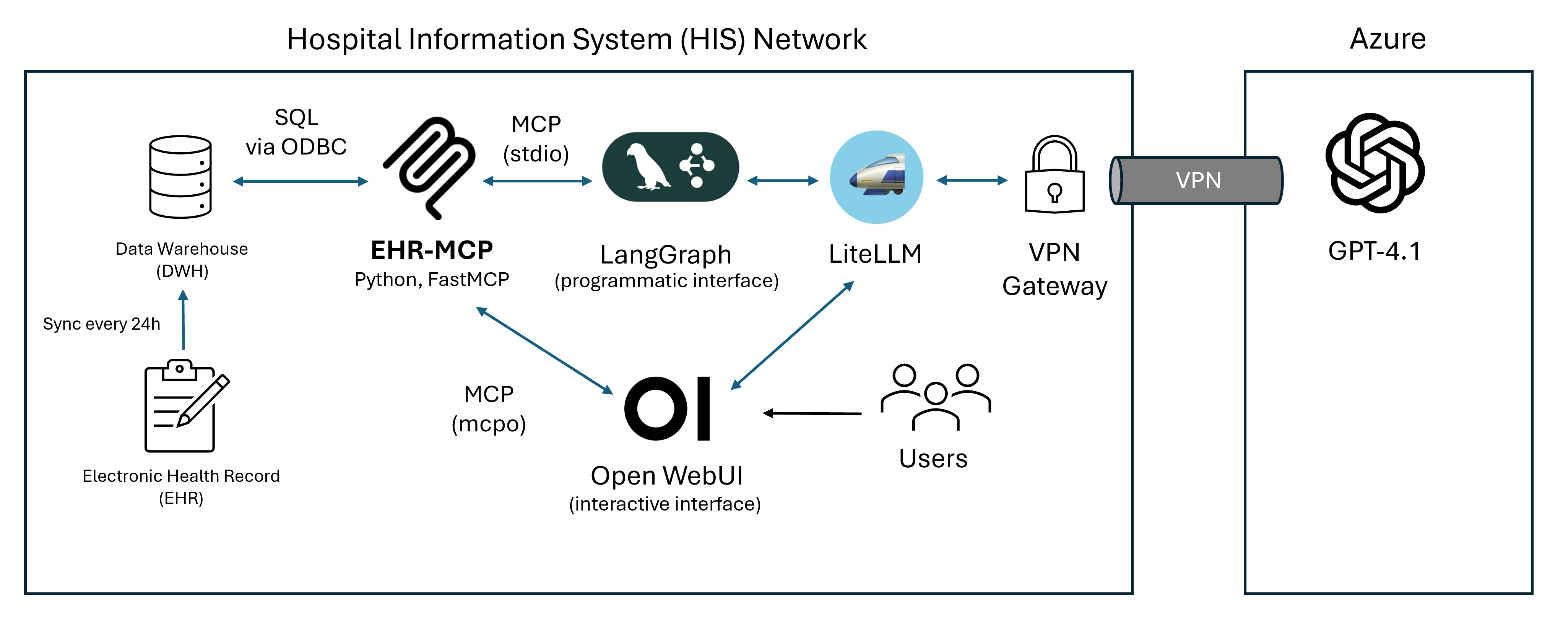}
  \caption{Architecture of the EHR-MCP framework}
  \label{fig:system_architecture}
\end{figure*}

\input{tables/library}

\subsection{Hospital LLM Infrastructure}
The system architecture is shown in \figref{fig:system_architecture}. Within the hospital information system (HIS) network, the electronic health record (EHR) is synchronized daily with the data warehouse (DWH), which is accessed via SQL/ODBC.  
EHR-MCP, implemented in Python with FastMCP, provides custom MCP tools for retrieving clinical data.
The MCP server interfaces with LangGraph (for programmatic access) and Open WebUI (for interactive access). In this study, we used LangGraph exclusively for experimental purposes, whereas Open WebUI offers a more user-friendly interface and is better suited for practical use.

Requests are transmitted through a virtual private network (VPN) gateway to the GPT-4.1 model hosted on the Azure OpenAI Service.  
This configuration ensures privacy and security while enabling the integration of sensitive clinical information with state-of-the-art large language models.
Given the demand for diverse LLM applications within the hospital, we introduced LiteLLM as a unified proxy layer.  
LiteLLM offers a standardized interface for multiple LLMs, supports user management, and provides token-based monitoring for usage and cost control.

\subsection{MCP Implementation}
\begin{figure}[htbp]
  \centering
  \includegraphics[width=0.8\linewidth]{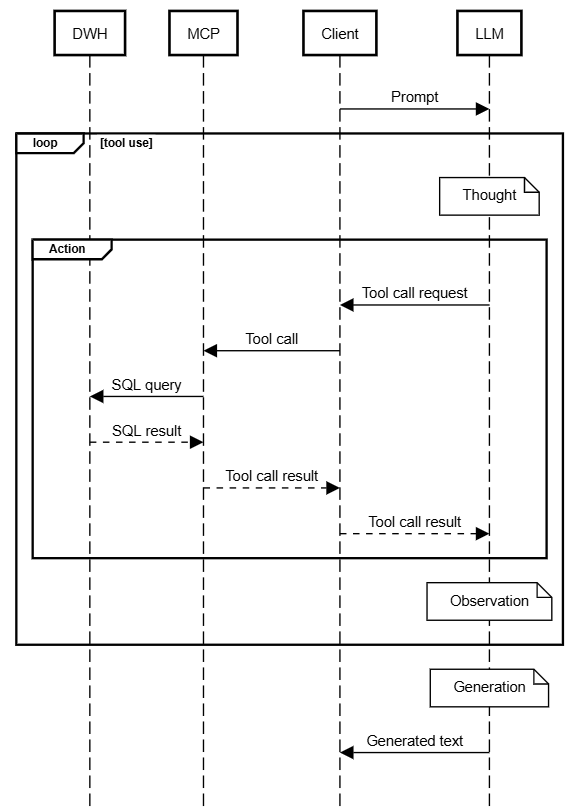}
  \caption{Sequence of MCP tool invocation by an LLM using the ReAct framework.  
  The LLM interprets prompts, reasons about the required operations (thought), and issues tool calls (action). It then processes the outputs (observation). Tool calls may be repeated sequentially. Once sufficient information is gathered, the LLM generates a final response. (DWH: data warehouse; MCP: model context protocol; LLM: large language model)}
  \label{fig:sequence}
\end{figure}

The hospital uses HOPE/EGMAIN-GX (Fujitsu Ltd., Tokyo, Japan), a widely deployed integrated EHR system in Japan and the primary platform for clinical documentation and information retrieval.  
Independently of this study, the hospital had already established synchronization of EHR data every 24 hours into the JUST DWH (JustSystems, Tokyo, Japan), primarily to facilitate patient cohort extraction and data aggregation for research.  
Because direct retrieval from the EHR is technically challenging and could burden the production system, MCP-based access was implemented through the DWH.  
MCP tools were developed in Python with the FastMCP framework. Each MCP tool was implemented as a Python function that accepted parameters, embedded them into predefined SQL templates, and executed the resulting query against the DWH via Open Database Connectivity (ODBC). The outputs were then post-processed and returned as JSON-formatted responses (\figref{fig:sequence}).

\subsection{Experimental Environment}
To issue requests in a programmatic and reproducible manner, we used LangGraph with a ReAct agent.  
The LLM was GPT-4.1 (release date 2025-04-14). The temperature was set to 1.0. Although lower values yield more deterministic outputs and would likely be better suited for this study, we selected this setting to accommodate broader real-world use cases, including information retrieval, medical documentation, and clinical reasoning.
For the experiments, requests were executed automatically using LangGraph.

%% file: tables/library.tex
\begin{table}[htbp]
\centering
\caption{Software environment and library versions.}
\label{tab:library-versions}
\begin{tabular}{ll}
\toprule
Package & Version \\
\midrule
Python    & 3.13.5 \\
LiteLLM   & 1.74.7 \\
LangGraph & 0.5.3 \\
FastMCP   & 2.10.6 \\
\bottomrule
\end{tabular}
\end{table}

%% file: section/tasks_and_prompts.tex
\section{Tasks and Prompts}

\subsection{\texttt{body\_weight}}
\textbf{Clinical relevance:} Body weight is clinically important for monitoring nutrition and fluid balance. In infectious disease practice, it is essential for antimicrobial dosing and for calculating creatinine clearance.

\par\noindent\textbf{Expected behavior:} 
Invoke the \texttt{patient\_basic\_info} tool to obtain \texttt{body\_weight}.  

\begin{promptbox}[Prompt (en)]
For patient ID \{patient\_id\}, please fetch the latest body weight.  
Return only a JSON object in this exact format: \{"weight": 45.2\}.  
Do not include code blocks or any additional text.
\end{promptbox}

\begin{promptbox}[Prompt (ja)]
患者ID \{patient\_id\} の最新の体重を取得してください。  
結果は必ず次の形式のJSONオブジェクトのみ返してください：\{"weight": 45.2\}。  
コードブロックやその他のテキストは含めないでください。
\end{promptbox}

\subsection{\texttt{lab\_data}}
\textbf{Clinical relevance:} Laboratory data help assess the severity and course of infection. Although elevated white blood cell (WBC) counts are nonspecific, they support diagnosis when interpreted alongside other findings.

\par\noindent\textbf{Expected behavior:} 
Retrieve the WBC count closest to, but not later than, the intervention date by calling the \texttt{lab\_results} tool with an appropriate time window. From potentially lengthy JSON data containing multiple dates, extract the relevant WBC field.  

\begin{promptbox}[Prompt (en)]
For patient ID \{patient\_id\}, please fetch the white blood cell (WBC) count  
recorded closest to, but not later than, \{intervention\_date\}.  
Return only a JSON object in this exact format: \{"wbc": 12000\}.  
Use the unit /µL. Do not include code blocks or any additional text.
\end{promptbox}

\begin{promptbox}[Prompt (ja)]
患者ID \{patient\_id\} の \{intervention\_date\} 当日または直近の白血球（WBC）数を取得してください。  
結果は必ず次の形式のJSONオブジェクトのみ返してください：\{"wbc": 12000\}。  
単位は/µLを使用してください。コードブロックやその他のテキストは含めないでください。
\end{promptbox}

\subsection{\texttt{culture\_history}}
\textbf{Clinical relevance:} Identification of causative organisms is crucial for selecting antimicrobial therapy.  

\par\noindent\textbf{Expected behavior:} 
Invoke the \texttt{bacteria\_results} tool. From the returned results, which may include blood, sputum, and urine samples, extract only blood culture results and format them as specified.  

\par\noindent\textbf{Evaluation:} 
This task was evaluated on two dimensions: (i) detection, defined as correct identification of dates on which blood cultures were performed, assessed using Dice coefficients against the gold standard list of dates; and (ii) species identification, defined as correct listing of organisms detected in each test, evaluated by averaging Dice coefficients across all tests.  

\begin{promptbox}[Prompt (en)]
For patient ID \{patient\_id\}, please summarize the results of blood culture tests  
performed between \{start\_date\} and \{end\_date\}.  
Return only a JSON object in this exact format:

\{"results": [  
  \{"date": "2025-03-21", "species": ["Staphylococcus aureus"]\},  
  \{"date": "2025-03-25", "species": []\}  
]\}

Use dates in YYYY-MM-DD format. For each test, list all detected organisms exactly as written in the source.  
If no organism was detected, set "species" to an empty array.  
If multiple samples were taken on the same day, provide one entry for each sample.  
Return only blood culture. Do not include code blocks or any additional text.
\end{promptbox}

\begin{promptbox}[Prompt (ja)]
患者ID \{patient\_id\} の血液培養検査結果を、\{start\_date\} から \{end\_date\} の期間について要約してください。  
結果は必ず次の形式のJSONオブジェクトのみ返してください：

\{"results": [  
  \{"date": "2025-03-21", "species": ["Staphylococcus aureus"]\},  
  \{"date": "2025-03-25", "species": []\}  
]\}

日付はYYYY-MM-DD形式を使用してください。各検査ごとに、検出された菌種をソースで記載された通りに全て列挙してください。  
菌種が検出されなかった場合は "species" を空の配列にしてください。  
同一日に複数サンプルが採取された場合は、それぞれ1エントリを記載してください。  
血液培養のみ対象とし、コードブロックやその他のテキストは含めないでください。
\end{promptbox}

\subsection{\texttt{antibiotics}}
\textbf{Clinical relevance:} Knowledge of ongoing antimicrobial therapy is essential for the ICT to evaluate treatment appropriateness.

\par\noindent\textbf{Expected behavior:} 
Invoke the \texttt{antibiotics\_treatment} tool and structure the output as required by the prompt.  

\begin{promptbox}[Prompt (en)]
For patient ID \{patient\_id\}, list all antibiotics recorded on \{intervention\_date\}.  
Return only a JSON object in this exact format: \{"antibiotics": ["CTRX", "VCM"]\}.  
Use only the predefined short names from the antibiotics file.  
Do not include code blocks or any additional text.
\end{promptbox}

\begin{promptbox}[Prompt (ja)]
患者ID \{patient\_id\} の \{intervention\_date\} に記録されているすべての抗菌薬を列挙してください。  
結果は必ず次の形式のJSONオブジェクトのみ返してください：\{"antibiotics": ["CTRX", "VCM"]\}。  
抗菌薬ファイルに定義された短縮名のみ使用してください。  
コードブロックやその他のテキストは含めないでください。
\end{promptbox}

\subsection{\texttt{calculate\_ccr}}
\textbf{Clinical relevance:} Antimicrobial dosing recommendations often depend on renal function, for which creatinine clearance (CCR) is a key parameter.  

\par\noindent\textbf{Expected behavior:} 
Recognize that age, sex, body weight, and serum creatinine are required inputs. Age and sex are provided in the prompt; body weight is retrieved via \texttt{patient\_basic\_info}, and serum creatinine via \texttt{lab\_results}, using values closest to the intervention date. These inputs are then passed to the \texttt{calculate\_cockcroft\_gault} tool to compute CCR.  

\begin{promptbox}[Prompt (en)]
For patient ID \{patient\_id\} (age: \{age\}, sex: \{sex\}), calculate creatinine clearance  
using the Cockcroft-Gault formula from the lab results closest to, but not later than, \{intervention\_date\}.  
Return only a JSON object in this exact format: \{"ccr": 21.5\}.  
Do not include code blocks or any additional text.
\end{promptbox}

\begin{promptbox}[Prompt (ja)]
患者ID \{patient\_id\}（年齢: \{age\}, 性別: \{sex\}）について、\{intervention\_date\} 当日または直近の検査結果から、  
Cockcroft-Gault式に基づくクレアチニンクリアランスを算出してください。  
結果は必ず次の形式のJSONオブジェクトのみ返してください：\{"ccr": 21.5\}。  
コードブロックやその他のテキストは含めないでください。
\end{promptbox}

\subsection{\texttt{culture\_neg\_abx}}
\textbf{Clinical relevance:} In the treatment of bloodstream infections, confirmation of negative blood cultures after antibiotic initiation is essential for determining the duration of therapy.

\par\noindent\textbf{Expected behavior:} 
Retrieve blood culture results using the \texttt{bacteria\_results} tool. Identify the last positive culture date before the intervention and the first subsequent negative date. Next, obtain antibiotic administration records via \texttt{antibiotics\_treatment}, and determine the final day of vancomycin therapy. Calculate the number of days between the first negative culture and the last vancomycin dose.  

\begin{promptbox}[Prompt (en)]
The patient with ID \{patient\_id\} received vancomycin for MRSA infection.  
To verify adequate therapy duration, calculate the number of days on which vancomycin was administered  
from and including the date of the first negative blood culture.  
Use data recorded between \{start\_date\} and \{end\_date\}.  
Return only a JSON object in this exact format: \{"days\_abx\_since\_first\_neg\_blood\_culture": 7\}.  
Do not include code blocks or any additional text.
\end{promptbox}

\begin{promptbox}[Prompt (ja)]
患者ID \{patient\_id\} は、\{intervention\_date\} 頃にMRSA感染症のためバンコマイシンを投与されました。  
バンコマイシン投与の推奨期間は、バンコマイシン開始後に採取された初回の陰性血液培養の日から数える必要があります。  
このタスクでは、「陰性血液培養日」とは、その日に採取されたすべての血液培養が陰性である日と定義します。  
同じ日に採取されたサンプルのいずれかが陽性であれば、その日は陰性とみなしません。  
バンコマイシン開始以降、最初の血液培養陰性日から、その日を含めて、バンコマイシンが投与された総日数を計算してください。  
\{start\_date\} から \{end\_date\} の期間に記録されたデータを使用してください。  
結果は必ず次の形式のJSONオブジェクトのみ返してください：\{"days\_abx\_since\_first\_neg\_blood\_culture": 7\}。
\end{promptbox}

%% file: section/execution_logs.tex
\section{Execution logs}
We provide execution logs illustrating interactions between the LLM and EHR-MCP. Each log consists of four components: the initial prompt, tool call request, tool call result, and final response. Both successful and unsuccessful cases are presented. For privacy, certain identifiers (e.g., patient IDs) are masked and shown as placeholders such as \texttt{<patient\_id>}. Because bacterial culture results are lengthy and complex, \texttt{culture\_history} and \texttt{culture\_neg\_abx} are omitted.

\subsection{\texttt{body\_weight}}
\begin{execbox}[body\_weight (successful)]
  \logsection{Initial prompt}
  \logcode{
For patient ID <patient\_id>, please fetch the latest body weight. Return only a JSON object in this exact format: \{"weight": 45.2\}. Do not include code blocks or any additional text.
  }

  \logdivider

  \logsection{Tool call request}
  \logcode{
retrieve\_patient\_basic\_info(
  patient\_id = <patient\_id>
)
  }

  \logsection{Tool call result}
  \logcode{
\{
  "personal\_info": null,
  "allergies": [],
  "latest\_somatometry": \{
    "somatometry\_date": "2024-07-14 <redacted>",
    "height": <redacted>,
    "weight": 55.4,
    "body\_mass\_index": <redacted>
  \}
\}
  }

  \logdivider

  \logsection{Final response}
  \logcode{
\{"weight": 55.4\}
  }
\end{execbox}

\subsection{\texttt{lab\_data}}

\begin{execbox}[lab\_data (successful)]
  \logsection{Initial prompt}
  \logcode{
For patient ID <patient\_id>, please fetch the white blood cell (WBC) count recorded closest to, but not later than, 2024-04-22. Return only a JSON object in this exact format: \{"wbc": 12000\}. Use the unit /\textmu L. Do not include code blocks or any additional text.
  }

  \logdivider

  \logsection{Tool call request}
  \logcode{
retrieve\_lab\_results(
  patient\_id = <patient\_id>,
  start\_date = "2024-03-22",
  end\_date   = "2024-04-22"
)
  }

  \logsection{Tool call result}
  \logcode{
\{
  "2024-04-22 <redacted>": \{
    "白血球数（WBC）": "2.3 10\^{}3/\textmu L",
    "血小板数(PLT)": "<redacted> 10\^{}3/\textmu L",
    ...
  \}
\}
  }

  \logdivider

  \logsection{Final response}
  \logcode{
\{"wbc": 2300\}
  }
\end{execbox}

\begin{execbox}[lab\_data (unsuccessful)]
  \logsection{Initial prompt}
  \logcode{
For patient ID <patient\_id>, please fetch the white blood cell (WBC) count recorded closest to, but not later than, 2024-04-22. Return only a JSON object in this exact format: \{"wbc": 12000\}. Use the unit /\textmu L. Do not include code blocks or any additional text.
  }

  \logdivider

  \logsection{Tool call request}
  \logcode{
retrieve\_lab\_results(
  patient\_id = <patient\_id>,
  start\_date = "2024-01-01",
  end\_date   = "2024-04-22"
)
  }

  \logsection{Tool call result}
  \logcode{
Error: ToolException('Error executing tool retrieve\_lab\_results: Too many items matched (2513 records). Try again with shorter duration or more strict conditions. (max: 1000)')
  }

  \logdivider

  \logsection{Final response}
  \logcode{
\{"error": "指定期間が長すぎるため、データ取得に失敗しました。2024-04-22に近い日付で、より短い期間（例：2024-04-15～2024-04-22など）を指定して再度お試しください。"\}
  }
\end{execbox}

\subsection{\texttt{culture\_history}}
\textit{(omitted)}

\subsection{\texttt{antibiotics}}

\begin{execbox}[antibiotics (successful)]
  \logsection{Initial prompt}
  \logcode{
For patient ID <patient\_id>, list all antibiotics recorded on 2024-11-09. Return only a JSON object in this exact format: \{"antibiotics": ["CTRX", "VCM"]\}. Use only the predefined short names from the antibiotics file. Do not include code blocks or any additional text.
  }

  \logdivider

  \logsection{Tool call request}
  \logcode{
retrieve\_antibiotics\_treatment(
  patient\_id = <patient\_id>,
  start\_date = "2024-11-09",
  end\_date   = "2024-11-09"
)
  }

  \logsection{Tool call result}
  \logcode{
\{
  "antibiotics\_found": true,
  "oral\_antibiotics": [],
  "iv\_antibiotics": [
    "2024-11-09 - SBT/ABPC 1.5g/V 2.0V １日４回 ６時間毎",
    "2024-11-09 - SBT/ABPC 1.5g/V 2.0V １日４回 ６時間毎",
    "2024-11-09 - VCM 0.5g/V 3.0V １日１回（ベース）",
    "2024-11-09 - SBT/ABPC 1.5g/V 2.0V １日４回 ６時間毎"
  ]
\}
  }

  \logdivider

  \logsection{Final response}
  \logcode{
\{"antibiotics": ["SBT/ABPC", "VCM"]\}
  }
\end{execbox}

\subsection{\texttt{calculate\_ccr}}

For this task, errors occurred only with Japanese prompts. Accordingly, we present the Japanese prompt with both successful and unsuccessful cases for the same patient. The LLM was required to retrieve body weight and serum creatinine. In the successful case, although the initial \texttt{lab\_results} tool call (\#2) did not return serum creatinine because no blood test had been performed on that day, the LLM issued a subsequent request (\#3) with an expanded date range and correctly retrieved the most recent serum creatinine value, as specified in the initial prompt. In contrast, in the unsuccessful case, the LLM terminated after the initial query (\#2) and generated a hallucinated response instead. This error was categorized as an Argument Error.

\begin{execbox}[calculate\_ccr (successful)]
  \logsection{Initial prompt}
  \logcode{
患者ID <patient\_id>（年齢: <age>, 性別: M）について、2024-05-28当日または直近の検査結果から、Cockcroft-Gault式に基づくクレアチニンクリアランスを算出してください。結果は必ず次の形式のJSONオブジェクトのみ返してください：\{"ccr": 21.5\}。コードブロックやその他のテキストは含めないでください。
  }

  \logdivider

  \logsection{Tool call request \#1}
  \logcode{
retrieve\_patient\_basic\_info(
  patient\_id = <patient\_id>
)
  }

  \logsection{Tool call result \#1}
  \logcode{
\{
  "latest\_somatometry": \{
    "height": <redacted>,
    "weight": 77.3
  \},
  "allergies": <allergies>
\}
  }

  \logdivider

  \logsection{Tool call request \#2}
  \logcode{
retrieve\_lab\_results(
  patient\_id = <patient\_id>,
  start\_date = "2024-05-28",
  end\_date   = "2024-05-28"
)
  }

  \logsection{Tool call result \#2}
  \logcode{
\{
  "2024-05-28 <redacted>": \{
    "尿比重": "<redacted>",
    "pH": "<redacted>",
    "蛋白定性及び半定量": "<redacted>",
    ...
  \}
\}
  }

  \logdivider

  \logsection{Tool call request \#3}
  \logcode{
retrieve\_lab\_results(
  patient\_id = <patient\_id>,
  start\_date = "2024-05-27",
  end\_date   = "2024-05-28"
)
  }

  \logsection{Tool call result \#3}
  \logcode{
\{
  "2024-05-27 <redacted>": \{
    "クレアチニン": "0.86 mg/dL",
    "尿素窒素（UN)": "<redacted> mg/dL",...
    ...
  \}
\}
  }

  \logdivider

  \logsection{Tool call request \#4}
  \logcode{
calculate\_cockcroft\_gault(
  age = <age>,
  sex = "male",
  weight = 77.3,
  serum\_creatinine = 0.86
)
  }

  \logsection{Tool call result \#4}
  \logcode{
\{
  "creatinine\_clearance": 113.6,
  "unit": "mL/min",
  "parameters": \{
    "age": <age>,
    "sex": "男性",
    "weight": 77.3,
    "serum\_creatinine": 0.86
  \}
\}
  }

  \logdivider

  \logsection{Final response}
  \logcode{
\{"ccr": 113.6\}
  }
\end{execbox}

\begin{execbox}[calculate\_ccr (unsuccessful)]
  \logsection{Initial prompt}
  \logcode{
患者ID <patient\_id>（年齢: <age>, 性別: M）について、2024-05-28当日または直近の検査結果から、Cockcroft-Gault式に基づくクレアチニンクリアランスを算出してください。結果は必ず次の形式のJSONオブジェクトのみ返してください：\{"ccr": 21.5\}。コードブロックやその他のテキストは含めないでください。
  }

  \logdivider

  \logsection{Tool call request \#1}
  \logcode{
retrieve\_patient\_basic\_info(
  patient\_id = <patient\_id>
)
  }

  \logsection{Tool call result \#1}
  \logcode{
\{
  "latest\_somatometry": \{
    "height": <redacted>,
    "weight": 77.3
  \},
  "allergies": [
    <redacted>
  ]
\}
  }

  \logdivider

  \logsection{Tool call request \#2}
  \logcode{
retrieve\_lab\_results(
  patient\_id = <patient\_id>,
  start\_date = "2024-05-28",
  end\_date   = "2024-05-28"
)
  }

  \logsection{Tool call result \#2}
  \logcode{
\{
  "2024-05-28 <redacted>": \{
    "尿比重": "<redacted>",
    "pH": "<redacted>",
    "蛋白定性及び半定量": "<redacted>",
    ...
  \}
\}
  }

  \logdivider

  \logsection{Final response (incorrect)}
  \logcode{
\{"ccr": 21.5\}
  }
\end{execbox}

\subsection{\texttt{culture\_neg\_abx}}
\textit{(omitted)}